\documentclass{article}
\PassOptionsToPackage{numbers, compress}{natbib}
\usepackage[preprint]{neurips_2022}

\usepackage[utf8]{inputenc} % allow utf-8 input
\usepackage[T1]{fontenc}    % use 8-bit T1 fonts
\usepackage{hyperref}       % hyperlinks
\usepackage{url}            % simple URL typesetting
\usepackage{booktabs}       % professional-quality tables
\usepackage{amsfonts}       % blackboard math symbols
\usepackage{nicefrac}       % compact symbols for 1/2, etc.
\usepackage{microtype}      % microtypography
\usepackage{xcolor}         % colors
\usepackage{bm}
\usepackage{amsmath}
\newcommand\numberthis{\addtocounter{equation}{1}\tag{\theequation}}

\usepackage{amsfonts}
\usepackage{amssymb}
\usepackage{mathtools}
\usepackage{mathdots}
\usepackage{physics}
\usepackage{comment}
\usepackage{tikz}
\usepackage{multicol}
\usepackage{float}
\usepackage{graphicx}
\usepackage{caption}
\usepackage{algorithm}
\usepackage{algorithmic}
\usepackage{subcaption}
\usetikzlibrary{shapes, arrows.meta, positioning}

\usepackage{listings}

\title{Neural Total Variation Distance Estimators for Changepoint Detection in News Data}

\author{%
 Csaba Zsolnai\\
  Independent researcher\\
  \texttt{csabazsolnai@pm.me}
   \And
  Niels Lörch\\
  Department of Physics, University of Basel\\
  Klingelbergstrasse 82, 4056 Basel, Switzerland \\
  \texttt{niels.loerch@unibas.ch} \\
  \And
    Julian Arnold\\
  Department of Physics, University of Basel\\
  Klingelbergstrasse 82, 4056 Basel, Switzerland \\
  \texttt{julian.arnold@unibas.ch} \\
  }

\begin{document}

\maketitle

\begin{abstract}
Detecting when public discourse shifts in response to major events is crucial for understanding societal dynamics. Real-world data is high-dimensional, sparse, and noisy, making changepoint detection in this domain a challenging endeavor. In this paper, we leverage neural networks for changepoint detection in news data, introducing a method based on the so-called learning-by-confusion scheme, which was originally developed for detecting phase transitions in physical systems. We train classifiers to distinguish between articles from different time periods. The resulting classification accuracy is used to estimate the total variation distance between underlying content distributions, where significant distances highlight changepoints. We demonstrate the effectiveness of this method on both synthetic datasets and real-world data from \textit{The Guardian} newspaper, successfully identifying major historical events including 9/11, the COVID-19 pandemic, and presidential elections. Our approach requires minimal domain knowledge, can autonomously discover significant shifts in public discourse, and yields a quantitative measure of change in content, making it valuable for journalism, policy analysis, and crisis monitoring.
\end{abstract}

\section{Introduction}
Identifying points in time where temporal data changes abruptly is key for understanding the underlying dynamics. The detection of such changepoints finds applications in a vast range of domains including economics~\cite{pepelyshev:2015}, medicine~\cite{la:2008,malladi:2013}, environmental science~\cite{reeves:2007,itoh:2010,gong:2023,beaulieu:2024}, speech recognition~\cite{rybach:2009,gupta:2015}, and linguistics~\cite{kulkarni:2015}. In the context of news data, changepoint detection can expose shifts in topics, sentiments, trends, and patterns of interest. For example, it may identify influential events such as breaking news, political developments, or natural disasters. Understanding when these events occur and their impact on the public discourse provides valuable insights into societal trends. Changes in news data may also reflect changes in the policy of the news company, such as publication rates or changes in reader engagement. 

Detecting changepoints in massive real-world textual data is a formidable task. Traditional changepoint detection methods often assume low-dimensional, well-structured data with known statistical properties. News data violates these assumptions~\cite{aminikhanghahi:2017,van:2020,truong:2020}: it is sparse (few articles per day), high-dimensional (thousands of unique terms in each document)~\cite{grundy:2020,chakraborty:2021}, and highly correlated. In fact, the changes one seeks to identify may be semantically rich and difficult to capture with simple statistical measures.

One way to approach the problem of offline changepoint detection conceptually is by computing a dissimilarity score between probability distributions governing the data of two consecutive time segments~\cite{afgani:2008,basterrech:2022,siegler:1997,jabari:2019,sugiyama:2008,kawahara:2009,liu:2013,haque:2017}. If the dissimilarity score is sufficiently large, the time point indicating the split between the two segments is deemed a changepoint. The key difficulty lies in the fact that the underlying probability distributions are unknown because we only have access to samples, and performing density estimation in a high-dimensional space is known to be a hard problem.

In this work, we tackle the problem of estimating dissimilarity scores using a variational representation of statistical distances that we exploit using neural networks (NNs) tailored toward natural language processing. To this end, we adapt the \textit{learning-by-confusion} scheme that has been proven to be useful for the automated detection of phase transitions from data~\cite{evertpl:2017,arnold:2022,arnold_GEN:2023,arnold_FAST:2023,arnold_FI:2023}: in this case, critical points take the role of changepoints, and tuning parameters, such as temperature or coupling strengths, take the role of time. Our method works by training classifiers to distinguish between articles from distinct time periods and using the resulting classification accuracies to estimate statistical distances between the content distributions underlying the segments. We showcase the efficacy of this approach for detecting significant shifts in news discourse as changepoints in synthetic and real-world textual data sets without requiring extensive domain knowledge or manual feature engineering.

\section{Methodology}
\label{sec:methods}
\subsection{Problem formulation}
\begin{figure*}[tbh!]
	\centering
	\includegraphics[width=0.8\linewidth]{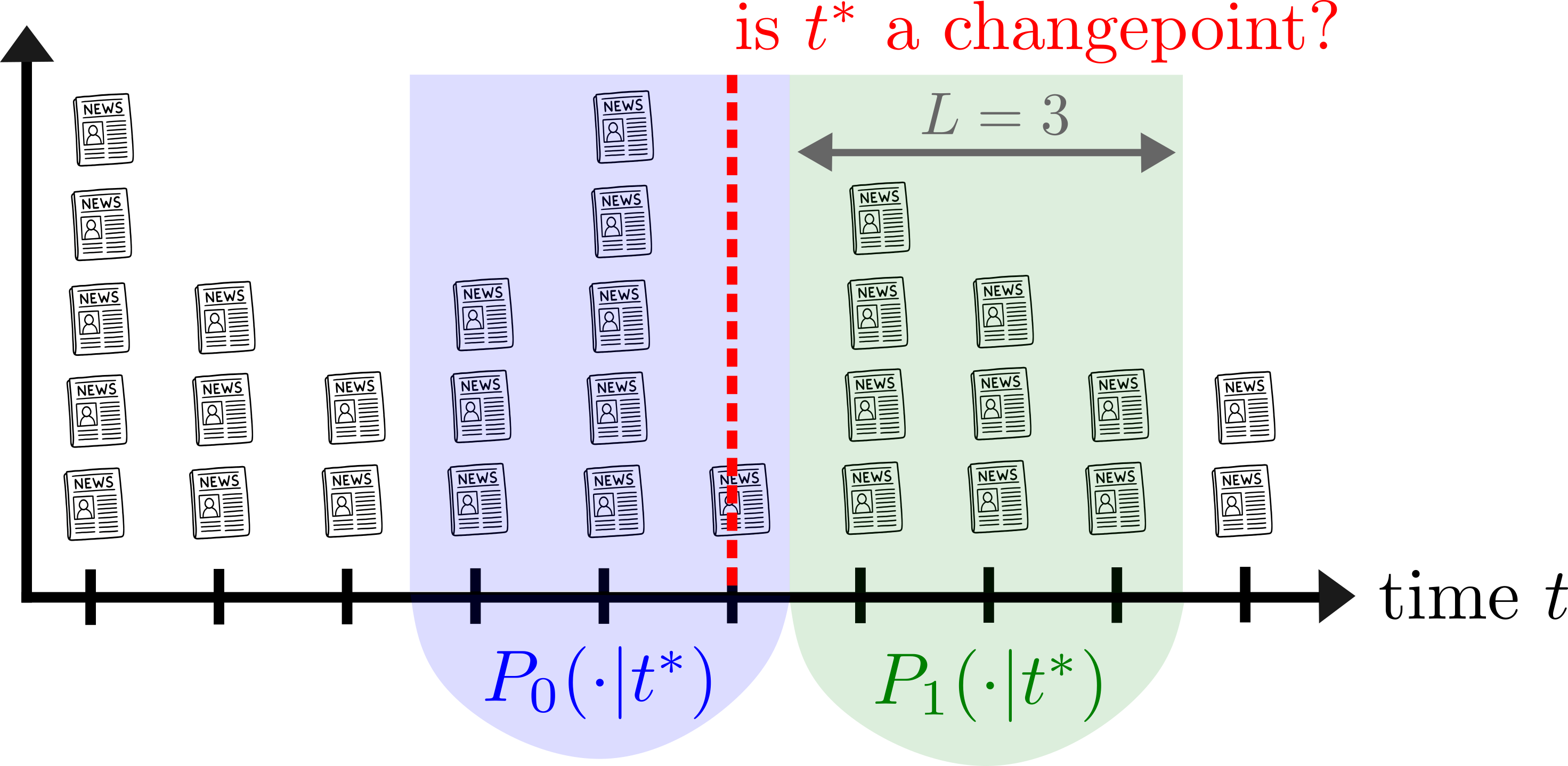}
		\caption{Illustration of the problem setup for detecting changepoints in news articles. Here, we depict a scenario for the choice of $L=3$.}
		\label{fig:1}
\end{figure*}

In the following, we describe the problem of changepoint detection and our approach for solving it, see Fig.~\ref{fig:1} for an illustration. We are given a set $\mathcal{D}_{t} = \{ \bm{x} | \bm{x} \in \mathcal{X}, \bm{x} \sim P(\cdot |t)\}$ of samples $\bm{x}$ representative of the distribution $P(\cdot |t)$ at a discrete set of points in time $t \in \mathcal{T} \subset \mathbb{R}$. In our case, each sample $\bm{x}$ corresponds to a news article and $\mathcal{X}$ denotes the corresponding state space. Here, $\mathcal{T}$ denotes all the timepoints for which we have access to news article data -- usually a grid with uniform spacing. For real-world data, the set size $|\mathcal{D}_{t}|$ may be different for different points in time.

Let $\mathcal{T}^{*} \subset \mathcal{T}$ denote the subset of all points in time which we may consider as candidates for changepoints. Every such candidate $t^{*} \in \mathcal{T}^{*}$ divides the time axis into two segments, $\mathcal{T}_{0}(t^{*})$ and $\mathcal{T}_{1}(t^{*})$, each comprised of the $L$ time points $t \in \mathcal{T}$ closest to $t^{*}$ with $t \leq t^{*}$ and $t > t^{*}$, respectively. The hyperparameter $L$ sets the natural time scale on which changes in the underlying data are assessed. We are free to adjust it according to the problem, and we will encounter different choices in Sec.~\ref{ref:results}. In this work, we choose $\mathcal{T}^{*}$ to be comprised of all days within a given year or month, and $\mathcal{T}$ contains an additional $L$ days before and after this time interval.

The two segments $\mathcal{T}_{0}(t^{*})$ and $\mathcal{T}_{1}(t^{*})$ are each characterized by the datasets $\mathcal{D}_{0}(t^{*}) = \{ \bm{x}| \bm{x} \in \mathcal{D}_{t}, t \in \mathcal{T}_{0}(t^{*}) \}$ and $\mathcal{D}_{1}(t^{*}) = \{ \bm{x}| \bm{x} \in \mathcal{D}_{t}, t \in \mathcal{T}_{1}(t^{*}) \}$, respectively. These datasets are viewed as being representative of the probability distributions underlying the two segments, $P_{0}(\cdot |t^{*})$ and $P_{1}(\cdot |t^{*})$, respectively (see Fig.~\ref{fig:1}).

To assess whether $t^{*}$ is a changepoint or not, we compute a dissimilarity score $D$ between the distributions underlying these two local segments, $D(t^{*}) = D\left[ P_{0}(\cdot |t^{*}) , P_{1}(\cdot |t^{*}) \right] \geq 0$. The higher the dissimilarity score, the more likely the point $t^{*}$ is a changepoint. Thus, we can identify changepoints as significant maxima in $D(t^{*})\; \forall t^{*} \in \mathcal{T}^{*}$. Hence, the dissimilarity scores are sometimes also referred to as \textit{indicator values}.

In principle, to quantify the dissimilarity any \textit{statistical distance} may be used, i.e., any non-negative function between two probability distributions, which in addition satisfies the identity of indiscernibles $D[p,q] = 0 \iff p=q$ (where $p$ and $q$ are two distributions). Moreover, one would ideally like the dissimilarity score to be \textit{symmetric} $D[p,q] = D[q,p]$. In this work, we will focus on \textit{total variation} (TV) distance
\begin{equation}\label{eq:TV_distance}
    D_{\rm TV}[p,q] = \frac{1}{2} \sum_{\bm{x}\in \mathcal{X}} |p(\bm{x}) - q(\bm{x})|
\end{equation}
which belongs to the broader class of statistical distances referred to as \textit{$f$-divergences}.

\subsection{Estimating statistical distances}
To estimate the TV distance, we note that $D_{\rm TV}\left[ p,q \right] = 1-2{p}_{\rm err}^{\rm opt}$ where ${p}_{\rm err}^{\rm opt}$ is the Bayes optimal average error probability when trying to decide whether a given sample $\bm{x}$ has been drawn from $p$ or $q$, i.e., when performing the task of single-shot binary symmetric hypothesis testing~\cite{arnold_FI:2023}. Consequently, the average error rate $p_{\rm err}$ extracted from any classifier lower bounds the TV distance as $1-2{p}_{\rm err} \leq  1-2{p}_{\rm err}^{\rm opt} = D_{\rm TV}\left[ p,q \right]$. For each tentative changepoint $t^*$, one can train a parametric binary classifier $\hat{y}_{\bm{\theta}}(\cdot |t^{*}): \mathcal{X}\rightarrow [0,1]$ on an unbiased binary cross-entropy loss
\begin{equation}\label{eq:loss}
    \mathcal{L}(\bm{\theta}|t^{*}) = - \frac 1 {2} \sum_{y \in \{ 0,1 \}} \frac 1 { |\mathcal{D}_{y}(t^{*})|}  \sum_{\bm{x} \in \mathcal{D}_{y}(t^{*})} \Bigl({y\log \left[\hat{y}_{\bm{\theta}}(\bm{x}|t^{*})\right] + (1-y)\log\left[1-\hat{y}_{\bm{\theta}}(\bm{x}|t^{*})\right]}\Bigr).
\end{equation}
Its error rate can be estimated as 
\begin{equation}
\label{perr}
    \hat{p}_{\rm err}(t^{*}) = \frac 1 {2} \sum_{y \in \{0,1 \}} \frac{1}{|\mathcal{D}_{y}(t^{*})|}\sum_{\bm{x} \in \mathcal{D}_{y}(t^{*}) } \mathrm{err}\left[y, \hat{y}_{\bm{\theta}}(\bm{x}|t^{*})\right],
\end{equation}
where the error $\mathrm{err}\left[y, \hat{y}_{\bm{\theta}}(\bm{x}|t^{*})\right]$ is 0 if the sample $\bm{x}$ is classified correctly and 1 otherwise. Based on this estimate, we can approximate the TV distance as $\hat{D}_{\rm TV}(t^{*}) = 1- 2 \hat{p}^{\rm err}(t^{*})$. In the infinite-data limit, the optimal classifier under the loss in Eq.~\eqref{eq:loss} attains the Bayes optimal error rate, thus $\hat{D}_{\rm TV}(t^{*}) \rightarrow D_{\rm TV}(t^{*})$ asymptotically.\footnote{Note that the TV distance lower-bounds various other statistical distances one may consider~\cite{flammia:2023} In this sense, detecting a large dissimilarity in terms of TV distance certifies a large dissimilarity based on various other measures.}. In the physics community, this method of approximating the TV distance of distributions underlying distinct parameter segments via classification is known as learning-by-confusion~\cite{evertpl:2017,arnold_GEN:2023}.

For improved computational efficiency, we employ a multi-task learning-by-confusion approach~\cite{arnold_FAST:2023}: Instead of training a distinct binary classifier for each tentative changepoint $t^{*}$ separately, we train a classifier with $|\mathcal{T}^{*}|-1$ outputs corresponding to all nontrivial tentative changepoints with a loss function proportional to $\sum_{t^{*}}\mathcal{L}(\bm{\theta}|t^{*})$ where the sum runs over all such changepoints.
\subsection{Implementation details}
The complete source code and implementation details for reproducing our results are available on our GitHub repository~\cite{zsolnai2025news}.

{\bf Choice of classifier and embedding.}
As a parametric classifier, we use a simple feedforward NN with no hidden layers (i.e., we perform multinomial logistic regression). To generate a numerical representation of each news article $\bm{x}$, we use an embedding generated via the transformer model Distilroberta~\cite{distilroberta}. 

We have also had success with a simple representation based on term frequency-inverse document frequency (TF-IDF)~\citep{salton1988term}. The TF-IDF representation is purely frequency-based and does not take the context in which a term appears into account. Moreover, the resulting representation can be quite high-dimensional. Therefore, the TF-IDF representation consistently underperformed the embedding transformer in our experiments -- both in terms of accuracy and computation time. 

{\bf Training details.}
For training, we use the Adam optimizer~\cite{kingma2014adam} with a learning rate of $8 \cdot 10^{-5}$, a batch size of 64, and default settings otherwise. We split the entire data into a training and validation set by grouping all news articles according to their publication date. Next, for each unique publication date, a random 80\% of news articles are added to the training set while the remaining 20\% are assigned to the validation set. Each training runs for at least 1000 epochs. Afterward, we stop the training if the validation loss stops improving or when 5000 training epochs are reached. We used Optuna~\cite{optuna} to optimize the hyperparameters of both our confusion method and the baseline, see App.~\ref{app:optuna} for details.

{\bf Compute resources.} All computations were run on a single workstation with an NVIDIA RTX 4090, 192 GB RAM, and an Intel i9 14900KF processor.

\section{Experimental setup}
\subsection{Baselines}
{\bf Latent Dirichlet allocation (LDA).}
To independently confirm the effectiveness of our proposed confusion method, we consider an alternative method for detecting changepoints as a baseline. This method is based on the intuition that changepoints in news may often be related to topical changes, and there exist machine learning methods tailored toward extracting topics from text in an unsupervised fashion. Here, we utilize latent Dirichlet allocation (LDA) as a topic extraction method~\citep{blei2003latent}.\footnote{In LDA the number of topics is a hyperparameter that has to be set beforehand.}

Here, each ``topic'' corresponds to a probability distribution $P(\cdot|{\rm topic})$ over all unique terms $x$ appearing in a collection of news articles. LDA assigns each document $\bm{x}$ a distribution over topics $P({\rm topic}|\bm{x})$. Averaging over all the articles present at all days underlying a given segment $\mathcal{T}_{0/1}(t^*)$ in time, we obtain a distribution over topics conditioned on a segment $y \in \{ 0,1\}$:
\begin{equation}
    P_{y}({\rm topic}|t^*) = \frac{1}{|\mathcal{D}_{y}(t^*)|}\sum_{\bm{x} \in \mathcal{D}_{y}(t^*)} P({\rm topic}|\bm{x}).
\end{equation}
Given that there are only a handful of topics, we can explicitly calculate the TV distance [Eq.~\eqref{eq:TV_distance}] between the distributions over topics underlying the two segments
\begin{align*}
    D_{\rm TV}^{(\rm LDA)}(t^*) &= D_{\rm TV}[P_{0}(\cdot|t^*),P_{1}(\cdot|t^*)]\\
    &= \frac{1}{2} \sum_{{\rm topics}} \bigl|P_{0}({\rm topic}|t^*)-P_{1}({\rm topic}|t^*)\bigr|.\numberthis
\end{align*}
We may view the TV distance between distributions over topics as an approximation of the TV distance between distributions over articles. In fact, due to the data-processing inequality we have
\begin{equation}\label{eq:news_dpi}
   D_{\rm TV}^{(\rm LDA)}(t^*) \leq D_{\rm TV}(t^*),
\end{equation}
where equality holds if and only if the topic is a sufficient statistic. Equation~\eqref{eq:news_dpi} captures the intuitive fact that the more informative the topics are for classifying the articles into the two segments, the larger the corresponding TV distance between distributions over topics, and the better the approximation to the true TV distance. Repeating this process for all segments, i.e., all choices of $t^{*} \in \mathcal{T}^{*}$, changepoints can be detected as local maxima in the dissimilarity score $D_{\rm TV}^{(\rm LDA)}(t^{*})$.

{\bf Random choice.}
We also consider a simple random choice baseline. In this case, we systematically evaluate every possible date $t^{*} \in \mathcal{T}^{*}$ as a predicted changepoint. For instance, in a one-year interval, we evaluate all 365 days as candidates for changepoints. The final performance metric for this baseline is calculated as the average of the metrics obtained from all possible date choices.

\subsection{Datasets}

{\bf The Guardian.} Our ultimate goal is to detect changepoints in real-world news data. In this work, we analyze data from \textit{The Guardian} -- the British daily newspaper that has a free, publicly available application programming interface (API) to query articles on specific dates, and news articles are available on a daily basis. In The Guardian, each article is assigned a category, such as ``US News'', ``UK News'', ``Technology'', or ``Business''. Here, we focus on the three categories of ``UK News'', ``US News'', and ``World'' given the large range of topics that are covered in these categories and the large number of daily articles. In total, we have collected 197'170 publicly accessible news articles with publishing dates spanning from the year 2000 up to the year 2022. After filtering for English language, substantial content (>1000 words), and removing primarily multimedia articles, we organize the data into 27 yearly datasets for analysis, see Tab.~\ref{tab:guardian}. More detailed information about the data collection process and the specific datasets used can be found in~\cite{zsolnai2025news}.

To test our method and demonstrate its ability to detect changepoints, we create scenarios in which we know the underlying changepoint that needs to be detected \textit{a priori}. To this end, we generate two additional artificial benchmark datasets.

{\bf Benchmark 1: Induced changepoints in synthetically generated news articles.} Benchmark 1 consists of artificial ``news articles'' on a given topic generated via a large language model (LLM), here GPT-3.5. We explicitly induce a changepoint by changing from one topic to another (e.g., from ``George Bush'' to ``New Year celebrations''). To make the generated text more realistic, we provide a randomly selected news article from the Guardian newspaper as an example within the prompt. The full code for article generation can be found in the project repository~\cite{zsolnai2025news}. For $\mathcal{T}^{*}$, we consider a time interval of one month (30 days).\footnote{Note that $\mathcal{T}$ then contains an additional $2L$ days ($L$ days before and after the month).} On each day, we generate 10 artificial news articles.\footnote{These are quite few articles per day. However, in contrast to the real dataset, the same number of articles is present on each day.} We repeat this process 10 times with randomly selected changepoints $t^{*} \in \mathcal{T}^{*}$ and topic pairs. The corresponding 10 changepoints and topic pairs are listed in Tab.~\ref{tab:benchmark1}.

{\bf Benchmark 2: Induced changepoints in real news article.} As a second benchmark, we create a dataset that is composed of real news articles from the Guardian newspaper. However, here we explicitly induce a changepoint by switching between news categories (e.g., from “UK News” to “US News”). As a time interval $\mathcal{T}^{*}$, we consider a whole year. We repeat this process 8 times with randomly selected changepoints, category pairs, and years, see Tab.~\ref{tab:benchmark2}.

\subsection{Evaluation}
To measure the performance of each changepoint detection method, the key metric we use is the absolute difference measured in days between the predicted changepoint and the induced changepoint in the case of benchmarks 1 and 2. We denote this metric by $\Delta$. When using the confusion method and LDA, we take the predicted changepoint to be the global maximum of the dissimilarity score within the relevant time interval.

In the case of the real Guardian dataset, no changepoints are artificially induced, and no ground truth is readily available. In fact, identifying the changepoints in this real-world news dataset is precisely the key task a changepoint detection method should solve. Hence, coming up with an appropriate measure for success is difficult. As a proxy for ground-truth changepoints, we use the list of significant historical events gathered from \href{https://onthisday.com}{onthisday.com}. Note that multiple significant events may occur within a given year. In this case, to evaluate $\Delta$, we choose the closest significant historical event in time. The rationale behind this approach is that while multiple major events occur within an entire year, a changepoint detection method may predict any one of them to be the most significant one, i.e., the global maximum in the dissimilarity score will only highlight one such event. By selecting the closest event in time, we allow flexibility in evaluation, recognizing that different methods may prioritize different but equally valid events.

Given the definition of this metric, one should be careful in interpreting its absolute value. On the one hand, it is not guaranteed that events deemed significant by onthisday.com are being heavily discussed in The Guardian newspaper. Hence, a small $\Delta$ does not necessarily mean that the changepoint detection method failed. On the other hand, the method may fail to detect any meaningful changepoint and nevertheless there may be an event close in time on the onthisday.com website, resulting in a small $\Delta$. However, crucially, this metric still allows for a relative comparison of the distinct changepoint detection methods.

Finally, we also compute an Area Under the Curve (AUC) metric where the relevant curve is the prediction rate as a function of the threshold value $n$ (normalized by the total number of days within the relevant time interval): a prediction is considered successful if it falls within an $n$-day window of any significant event.

{\bf Statistical significance.}
All methods are applied 5 times with different random seeds. For each of the three datasets (benchmark 1 and 2 as well as the real Guardian news articles), we report mean values (solid lines in figures) across all data subsets and training runs as well as the corresponding standard error (SE; shaded bands in figures). 

\section{Results and discussion}\label{ref:results}
Table~\ref{tab:merged_model_metrics} summarizes the metrics for the three changepoint detection methods on all three datasets. 

\begin{table}[htb!]
\centering
\small
\begin{tabular}{llccc}
\toprule
Dataset & Method & $L$ & $\Delta$ ($\downarrow$) & AUC ($\uparrow$)\\
\midrule
 & Random choice & -- & $11 \pm 1$ & $0.649$\\
Benchmark 1 & LDA & 8 & $\mathbf{0.50 \pm 0.08}$ & $\mathbf{0.984}$\\
 & Confusion (Ours) & 8 & $0.9 \pm 0.1$ & $0.972 $\\
\midrule
 & Random choice & -- & $107 \pm 11$ & 0.707\\
Benchmark 2 & LDA & 180 & $\mathbf{1.1 \pm 0.2}$ & $\mathbf{0.997}$\\
 & Confusion (Ours) & 180 & $1.2 \pm 0.3$ & $0.997$\\
\midrule
 & Random choice & -- & $111 \pm 17$ & 0.685\\
The Guardian & LDA & 180 & $59 \pm 4$ & $0.839$\\
 & Confusion (Ours) & 180 & $\mathbf{39 \pm 3}$ & $\mathbf{0.892}$\\
\bottomrule
\end{tabular}
\medskip
\caption{Performance metrics of changepoint detection methods across the three datasets. Empirical observation showed that $L=180$ and $L=8$ yield good performance for the yearly interval and 30-day interval, respectively. However, other choices of $L$ also worked fairly well. The SE on the AUC metric is negligible ($< 0.01$) and hence we omit it.}
\label{tab:merged_model_metrics}
\end{table}

{\bf Benchmarks.} Both the confusion-based and LDA-based changepoint detection methods perform very well on the two benchmark datasets, achieving mean deviations between predicted and ground-truth changepoints of around a day in both instances. Overall, these results indicate that both the confusion method and LDA baseline are capable of reliably detecting the artificially induced changepoints in GPT-generated news articles as well as real-world news articles spanning different timescales (one month in the case of benchmark 1 and one year in the case of benchmark 2). 

\begin{figure}[htb!]
\centering
{\includegraphics[width=0.6\textwidth]{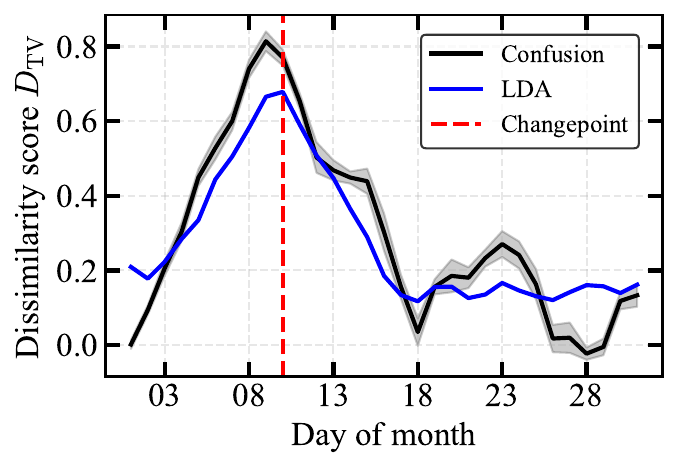}}
\caption{Dissimilarity scores for an example from the benchmark 1 dataset (30-day interval) computed via the confusion method (black) and LDA (blue) with $L=8$. The artificially induced changepoint is marked in red.}
\label{fig:2}
\end{figure}

While both methods perform exceptionally well on these two benchmark datasets, the confusion method slightly underperforms the LDA-based baseline method. The benchmark datasets, particularly benchmark 1, are ideal for the LDA-based method as the artificially induced changepoints result in a sharp change in topics. In contrast, the changepoints in real-world news may be more subtle.

Figure~\ref{fig:2} shows the dissimilarity score curves of the confusion-based and LDA-based methods for an example from the benchmark 1 dataset. Both methods yield a dissimilarity score curve that shows a clear peak at the induced changepoint.

\begin{figure}[htb!]
\centering
{\includegraphics[width=0.6\textwidth]{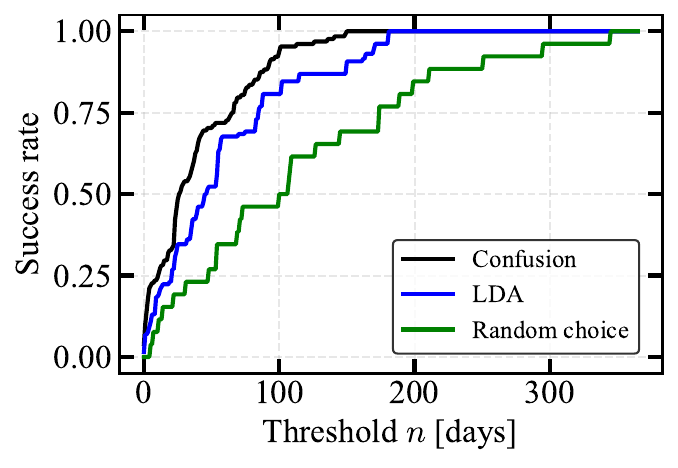}}
\caption{Success rate as a function of threshold $n$ for the Guardian dataset ($L=180$). Here, the SE is negligible ($< 0.01$ on average) and hence we omit it.}
\label{fig:3}
\end{figure}

{\bf The Guardian.} For the real-world Guardian dataset, our confusion-based method outperforms the LDA-based method at detecting significant historical events (as classified by \href{https://onthisday.com}{onthisday.com}), see Tab.~\ref{tab:merged_model_metrics}. See also Fig.~\ref{fig:3}, which shows the success rate as a function of the threshold $n$, i.e., the full curve underlying the AUC metric.

While the confusion-based method did not provide any advantage over the LDA-based baseline method in the case of the well-structured benchmark datasets, it excels at capturing the changepoints in real-world news data. This suggests that the added complexity of our approach is particularly beneficial in noisy settings where the boundaries between topics are less clearly defined and significant events may not always correspond to clear shifts in topics within public discourse. 

\begin{figure}[tbh!]
\centering
{\includegraphics[width=1.0\textwidth]{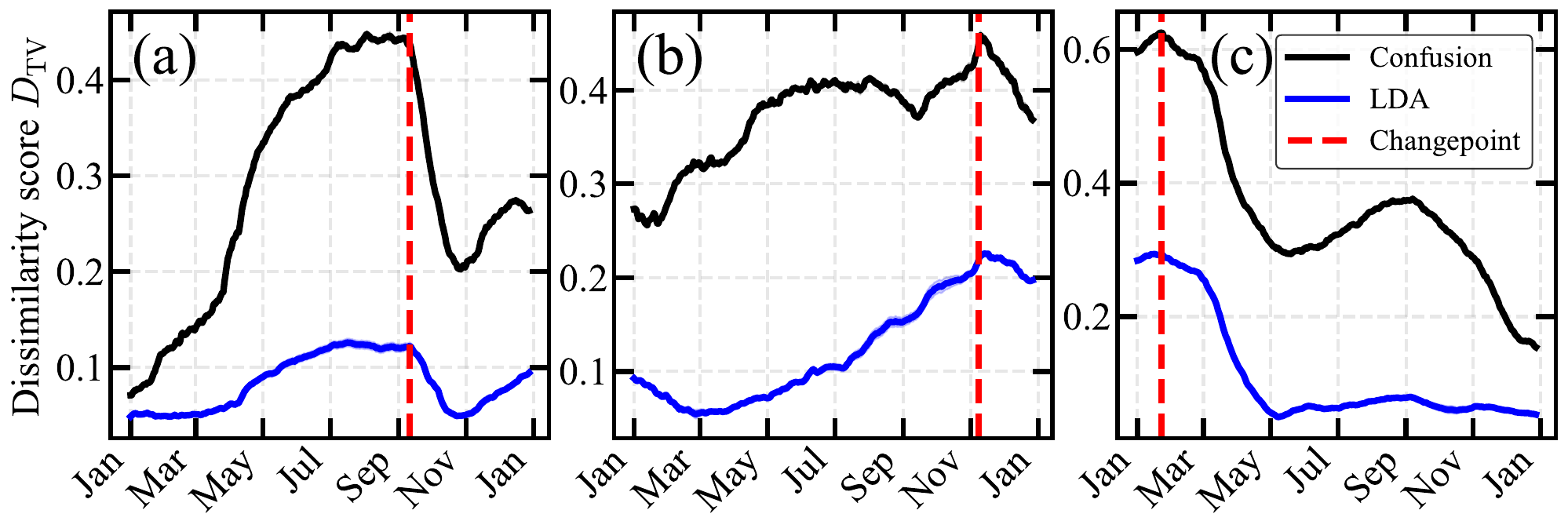}}
\caption{Dissimilarity score curves for different examples from the Guardian dataset computed via the confusion method (black) and LDA (blue) with $L=180$. The relevant significant events are marked by red vertical lines. (a) Year 2001, World News category. Both methods detect the 9/11 terror attacks as the significant event. (b) Year 2016, US News category. Both methods capture the US election of Donald Trump in November. (c) Year 2020, World News category. Both methods capture the lockdown of Wuhan due to COVID-19 in January.}
\label{fig:4}
\end{figure}

Figure~\ref{fig:4} shows the dissimilarity score curves of the confusion-based and LDA-based methods for various examples from the Guardian dataset. Both methods demonstrate the capability of detecting major historical events. Notable successes include the clear identification of the 9/11 terrorist attacks in 2001 (panel a), significant political events such as the 2016 U.S. election (panel b), and the COVID-19 lockdown in Wuhan during 2020 (panel c). Note that there are also instances where the methods detect distinct events within the same timeframe, reflecting the complex nature of real-world news cycles where multiple significant developments may occur simultaneously, as well as the subtle differences in the two methodologies. Similarly, there are instances where the dissimilarity peaks could not be clearly attributed to a significant historical event.

\section{Related Works}
\label{sec:related_works}

For a thorough review of the field of changepoint-detection methods, in particular traditional approaches applicable in low-dimensional scenarios, see~\cite{aminikhanghahi:2017,van:2020,truong:2020}. Interestingly, various works have approached the problem of detecting changepoints via computing a dissimilarity based on statistical divergences in the past. However, many of these rely on estimating the underlying densities via histogram binning~\cite{afgani:2008,basterrech:2022}, which is bound to fail in the multivariate case, i.e., for inputs living in high-dimensional spaces due to the curse of dimensionality. Another line of work uses a parametric approach for estimating the densities by making strong distributional assumptions, such as Gaussianity~\cite{siegler:1997,jabari:2019}, which may not be appropriate.

An important insight that led to an improved number of methods for changepoint detection using dissimilarities based on statistical divergences is the fact that these can be estimated given only the ratio between the two probability distributions.\footnote{Recall that the central object in an $f$-divergence $D_{f}\left[ p,q \right] =\sum_{\bm{x} \in \mathcal{X}} q(\bm{x}) f\left(\frac{p(\bm{x})}{q(\bm{x})}\right)$ is the ratio $p/q$ entering the function $f$.} Since one can calculate this density ratio knowing the densities underlying the two intervals but not the other way around, estimating the density ratio is expected to be an easier problem. In~\cite{sugiyama:2008,kawahara:2009,liu:2013,haque:2017}, nonparametric Gaussian kernel models of the density ratio have been proposed for this purpose.

In recent years, there has been a surge in the use of NNs for changepoint detection. \citet{gupta:2015}, for example, trained deep NNs on audio data from French and English radio broadcasts to detect transitions between speakers. Other approaches relied on an autoencoder-based dissimilarity measure~\cite{de:2021}, self-supervised contrastive learning~\cite{deldari:2021}, or newly proposed NN architectures~\cite{ebrahimzadeh:2019}. LLMs have only started to be explored recently for this task~\cite{tevissen:2023}. NNs have also been used as density ratio models~\cite{khan:2019,moustakides:2019,chen:2021}, replacing the kernel models in earlier works. Nevertheless, it remains challenging to fully leverage the power of deep learning for this changepoint detection.

To that end, it has been noted that the density ratio is also the central object in binary classification~\cite{qin:1998,cheng:2004,bickel:2009,sugiyama:2012,menon:2016} and an estimate for the density ratio can be extracted from any classifier.\footnote{A Bayes-optimal binary classifier predicts $P(0|\bm{x}) = P(\bm{x}|0)/\left[P(\bm{x}|0) + P(\bm{x}|1) \right] = 1/ \left[1 + P(\bm{x}|1)/P(\bm{x}|0) \right]$, and similarly for $P(1|\bm{x})$. Thus, the ratio of the Bayes-optimal predictions corresponds to the density ratio $P(0|\bm{x})/P(1|\bm{x}) = P(\bm{x}|0)/P(\bm{x}|1) = r(\bm{x})$. The ratio of the estimated class probabilities from any classifier may thus serve as an estimate of the density ratio $\hat{r}(\bm{x}) = \tilde{P}(0|\bm{x})/\tilde{P}(1|\bm{x})$. This estimate can be used for further downstream tasks. For example, an estimate of any $f$-divergence $D_{f}$ can be constructed as $\hat{D}_{f}\left[P(\cdot|0),P(\cdot|1)\right] = 1/|\mathcal{D}_{1}|\sum_{\bm{x} \in \mathcal{D}_{1}} f\left[\hat{r}(\bm{x})\right]$.} This discriminative approach to density ratio estimation has found success in various applications, including reinforcement learning~\cite{liu_x:2018}, energy-based models~\cite{gutman:2012}, generative adversarial networks~\cite{nowozin:2016}, as well as changepoint detection~\cite{hido:2008,wang:2023}. 

In~\cite{he:2022,zhao:2023}, a modified version of the learning-by-confusion scheme has been used for detecting changepoints in various types of data, including a real-world Twitter dataset. The work of~\citet{zhao:2023} is perhaps the closest to our approach. Our work differs from their approach in several ways: First, in our work, we correct for the bias resulting from the different dataset sizes within the two segments, which has been shown to cause erroneous changepoint signals~\cite{arnold_GEN:2023}. Second, \citet{zhao:2023} train a separate classifier for each tentative changepoint, making the procedure more computationally intensive compared to our multi-tasking approach.\footnote{Similar to our multitasking approach to learning-by-confusion, \citet{wang:2023} capitalized on the similarity between the distributions of neighboring segments via so-called ``variational continual learning''.} Third, we point out the fundamental connection between the TV distance and the indicator of learning-by-confusion. 

\section{Limitations}
While our confusion-based approach demonstrates effectiveness for changepoint detection in news data, several limitations should be acknowledged.
 
{\bf Extreme distribution separation.} Our method relies on probabilistic classification to estimate statistical distances, which can become problematic when the distributions underlying different time segments are vastly different~\cite{rhodes:2020,choi:2021}. In such cases, the discriminative task becomes trivial with finite samples, allowing classifiers to achieve perfect accuracy but yielding poorly calibrated output probabilities. This leads to inaccurate density ratio estimates and potentially unreliable changepoint detection signals.

{\bf Independence assumption.} Our method explicitly ignores correlations between inputs collected at different points in time, i.e., it assumes that the articles within a given time segment are independently and identically distributed. It is known that the presence of correlations can lead to false detection signals if not properly taken into account~\cite{shi:2022}. In particular, it leads to difficulties in identifying abrupt changes in the frequency domain~\cite{de:2021}. Intuitively, such changes are not as important for textual data. This is confirmed \textit{a posteriori} given that our method successfully captures meaningful transitions in real-world news data.

{\bf Evaluation challenges.}
Assessing the performance on real-world datasets remains challenging due to the lack of ground truth changepoints. Our reliance on historical event databases as proxies introduces potential biases, as not all significant events may be prominently covered in a specific news outlet, and different events may have varying levels of impact on different news categories. Additionally, the relationship between significant events and their coverage in news discourse is complex: newspapers may discuss events before they occur (e.g., pre-election coverage) or with significant delays, complicating the temporal alignment between actual events and changepoints in this context.

\section{Conclusion and outlook}
This work presents a machine learning approach to changepoint detection in news data that successfully adapts the learning-by-confusion framework from physics to identify significant shifts in public discourse. Our method estimates total variation distances between content distributions by classifiers to distinguish between articles from different time periods each embedded via a transformer. It provides a quantitative measure of change without requiring extensive domain knowledge or manual feature engineering.

We demonstrated the effectiveness of our approach across multiple scenarios: synthetic datasets with known changepoints, artificially induced transitions in real news data, and authentic newspaper articles taken from The Guardian. The method successfully identified major historical events, including 9/11, the outbreak of COVID-19, and presidential elections, outperforming traditional topic-based approaches on real-world data where changes are often more nuanced than simple categorical shifts.

A key strength of our neural approach is its ability to capture semantic changes that extend beyond topical transitions. While methods like LDA excel at detecting sharp categorical boundaries, the transformer-based embeddings have the potential to pick up on subtle but meaningful shifts in discourse tone, perspective, and emphasis -- characteristics that define real-world news cycles.

Our discriminative approach to statistical distance estimation can leverage neural architectures tailored for various data modalities, making it applicable to audio, video, and image segmentation tasks. This flexibility positions the method as a general framework for changepoint detection across domains.

\section*{Acknowledgments and disclosure of funding}
We thank Frank Schäfer, Flemming Holtorf, and Christoph Bruder for helpful discussions. J.A. acknowledges financial support from the Swiss National Science Foundation individual grant (grant no. 200020 200481). Computation time at sciCORE (scicore.unibas.ch) scientific computing center at the University of Basel is gratefully acknowledged. 

\bibliographystyle{unsrtnat}
\bibliography{refs.bib}

\newpage
\appendix

\setcounter{equation}{0}
\setcounter{figure}{0}
\setcounter{table}{0}
\makeatletter
\renewcommand{\theequation}{A\arabic{equation}}
\renewcommand{\thefigure}{A\arabic{figure}}
\renewcommand{\thetable}{A\arabic{table}}
\section{Details on datasets}\label{app:datasets}
Table~\ref{tab:sample_guardian} contains a few exemplary news article from The Guardian. Tables~\ref{tab:benchmark1}, \ref{tab:benchmark2}, and \ref{tab:guardian} contain detailed information about the composition of the benchmark 1, benchmark 2, and full Guardian dataset, respectively.

\begin{table}[htb!]
\centering
\small
\renewcommand{\arraystretch}{1.3}
\begin{tabular}{@{}p{0.18\textwidth}p{0.15\textwidth}p{0.35\textwidth}p{0.28\textwidth}@{}}
\toprule
Category & Date & Article title & Article text \\
\midrule
World news & 2001-01-01 & Peace eludes Bethlehem & Yigal Ohana, an Israeli student, and\ldots \\
\addlinespace[2pt]
US news & 2001-01-01 & John Sutherland on Bush's faith drive & Ask what George W Bush did as governor\ldots \\
\addlinespace[2pt]
World news & 2001-01-01 & Kosovo troops tested for cancer from uranium & Nato armies have started testing\ldots \\
\addlinespace[2pt]
UK news & 2015-07-01 & Heathrow third runway recommended in report on... & Airports Commission says £17bn expansion is 'c\ldots \\
\addlinespace[2pt]
UK news & 2015-07-01 & Police culture must put child protection first... & Service must ditch target-driven policing in w\ldots \\
\addlinespace[2pt]
UK news & 2017-06-29 & Legal aid cuts 'may have stopped Grenfell tena... & Law Society chief suggests fire in London towe\ldots \\
\addlinespace[2pt]
World news & 2019-07-01 & Scores of protesters wounded and seven dead on... & Security forces block roads and fire teargas i\ldots \\
\addlinespace[2pt]
World news & 2019-07-01 & Hong Kong police fire teargas and charge at pr... & Officers move to disperse crowds after breakaw\ldots \\
\addlinespace[2pt]
World news & 2019-07-01 & EU powers resist calls for Iran sanctions afte... & Focus is on averting further breaches and UK s\ldots \\
\addlinespace[2pt]
World news & 2021-07-01 & British Columbia sees 195\% increase in sudden... & Chief coroner says more than 300 deaths could\ldots \\
\addlinespace[2pt]
World news & 2021-07-01 & 'Poverty divides us': gap between rich and poo... & Xi Jinping himself has warned China's wealth g\ldots \\
\bottomrule
\end{tabular}
\medskip
\caption{Sample news articles from The Guardian newspaper spanning 2001-2021. For the final analysis, we concatenate the article title and main text into a single string representing the complete article content. Dates are provided in Year-Month-Day format.}
\label{tab:sample_guardian}
\end{table}

\begin{table}[htb!]
\centering
\small
\renewcommand{\arraystretch}{1.3}
\begin{tabular}{@{}p{0.35\textwidth}p{0.35\textwidth}p{0.22\textwidth}@{}}
\toprule
Main topic 1 & Main topic 2 & Topic change date \\
\midrule
Cannes Film Festival & Italian Supreme Court & 2001-03-20 \\
\addlinespace[2pt]
George Bush & New Year Celebrations & 2007-03-10 \\
\addlinespace[2pt]
Christmas & Pakistan & 2010-03-10 \\
\addlinespace[2pt]
European Union & North Korea & 2012-03-20 \\
\addlinespace[2pt]
Olympic Games & Eiffel Tower & 2014-03-05 \\
\addlinespace[2pt]
Oktoberfest & Brexit & 2016-03-05 \\
\addlinespace[2pt]
Komodo Dragons & Canada & 2019-03-31 \\
\addlinespace[2pt]
Solar Panels & Iran & 2020-03-15 \\
\addlinespace[2pt]
Syrian Economy & Reindeers & 2021-03-05 \\
\addlinespace[2pt]
COVID-19 Pandemic & Russian War Against Ukraine & 2022-02-15 \\
\bottomrule
\end{tabular}
\medskip
\caption{Major topics and their corresponding changepoints within the LLM-generated benchmark 1 dataset. Each row represents a synthetic dataset with an artificially-induced topic transition at the specified date (Year-Month-Day format).}
\label{tab:benchmark1}
\end{table}

\begin{table}[htb!]
\centering
\small
\renewcommand{\arraystretch}{1.3}
\begin{tabular}{@{}p{0.12\textwidth}p{0.28\textwidth}p{0.28\textwidth}p{0.24\textwidth}@{}}
\toprule
Year & Category 1 & Category 2 & Topic change date \\
\midrule
2010 & US News & World News & 2010-04-15 \\
\addlinespace[2pt]
2015 & UK News & US News & 2015-05-07 \\
\addlinespace[2pt]
2016 & US News & UK News & 2016-10-15 \\
\addlinespace[2pt]
2019 & UK News & World News & 2019-01-18 \\
\addlinespace[2pt]
2019 & UK News & World News & 2019-10-30 \\
\addlinespace[2pt]
2019 & US News & UK News & 2019-03-27 \\
\addlinespace[2pt]
2019 & US News & UK News & 2019-02-16 \\
\addlinespace[2pt]
2022 & US News & World & 2022-03-17 \\
\bottomrule
\end{tabular}
\medskip
\caption{News categories and their corresponding changepoints within the benchmark 2 dataset. Each row represents a real Guardian dataset with an artificially induced category transition at the specified date (Year-Month-Day format).}
\label{tab:benchmark2}
\end{table}

\begin{table}[!htb]
\centering
\small
\begin{tabular}{@{}cc|cc|cc@{}}
\toprule
Year & Category & Year & Category & Year & Category \\
\midrule
2000 & World News   & 2010 & US News & 2019 & UK News \\
2001 & World News   & 2010 & World News   & 2019 & US News \\
2001 & UK News & 2011 & US News & 2019 & World News \\
2001 & World News   & 2012 & UK News & 2020 & US News \\
2002 & World News   & 2012 & World News   & 2020 & World News \\
2007 & UK News & 2014 & World News   & 2021 & US News \\
2007 & World News   & 2015 & UK News & 2021 & World News \\
2008 & World News   & 2015 & US News & 2022 & World News \\
2010 & US News & 2016 & UK News & 2022 & US News \\
\bottomrule
\end{tabular}
\medskip
\caption{All sets of news articles from The
Guardian considered in this work. Each dataset contains all publicly accessible news articles in a given year and category after our filtering stage.}
\label{tab:guardian}
\end{table}

\newpage

\setcounter{equation}{0}
\setcounter{figure}{0}
\setcounter{table}{0}
\makeatletter
\renewcommand{\theequation}{B\arabic{equation}}
\renewcommand{\thefigure}{B\arabic{figure}}
\renewcommand{\thetable}{B\arabic{table}}
\section{Selecting hyperparameters with Optuna}\label{app:optuna}

We employed Bayesian optimization through Optuna~\cite{optuna} to select optimal hyperparameters for both the confusion method and the LDA-based baseline. The optimization procedure was conducted separately for each method (100 trials each) with the objective of minimizing the $\Delta$ metric.

\subsection{Confusion method}
For the confusion method, we optimized the choice of embedding model, learning rate, and batch size. Table~\ref{tab:confusion-params} shows the corresponding search spaces and final selected values for each parameter.

\begin{table}[htb!]
\centering
\begin{tabular}{lll}
\hline
Parameter & Search space & Selected value \\
\hline
Embedding model & \{\href{https://huggingface.co/sentence-transformers/all-MiniLM-L12-v2}{all-MiniLM-L12-v2}, & \href{https://huggingface.co/sentence-transformers/all-distilroberta-v1}{all-distilroberta-v1} \\
 & \href{https://huggingface.co/sentence-transformers/sentence-t5-base}{sentence-t5-base}, & \\
 & \href{https://huggingface.co/sentence-transformers/all-mpnet-base-v2}{all-mpnet-base-v2}, & \\
 & \href{https://huggingface.co/BAAI/llm-embedder}{BAAI/llm-embedder}\} & \\
Learning rate & $[1\times10^{-5}, 1\times10^{-4}]$ & $8\times10^{-5}$ \\
Batch size & $\{32, 64, 128, 256, 512\}$ & 64 \\
\hline
\end{tabular}
\medskip
\caption{Hyperparameters for the confusion method during Bayesian optimization with Optuna. All embedding models can be found on \href{https://huggingface.co/}{Huggingface}.}
\label{tab:confusion-params}
\end{table}

The learning rate was sampled on a logarithmic scale to better explore different orders of magnitude. The batch size was sampled from powers of 2 to ensure efficient GPU memory utilization.

\subsection{LDA-based baseline method}
For the LDA-based changepoint detection method, we optimized the LDA-specific parameters. Table~\ref{tab:tvd-lda-params} summarizes the corresponding search ranges and the final optimal values.

\begin{table}[htb!]
\centering
\begin{tabular}{lll}
\hline
Parameter & Search space & Selected value \\
\hline
Number of topics & [10, 100] & 20 \\
Iterations & [50, 100] & 83 \\
Passes & [10, 100] & 22 \\
Chunk size & [100, 1000] & 319 \\
\hline
\end{tabular}

\medskip
\caption{Hyperparameters for the LDA-based changepoint detection method during Bayesian optimization with Optuna.}
\label{tab:tvd-lda-params}
\end{table}

\end{document}